\documentclass{article}

    \usepackage[final, nonatbib]{neurips_2021}

\usepackage[utf8]{inputenc} 
\usepackage[T1]{fontenc}    
\usepackage{hyperref}       
\usepackage{url}            
\usepackage{booktabs}       
\usepackage{amsfonts}       
\usepackage{nicefrac}       
\usepackage{microtype}      
\usepackage{xcolor}         

\usepackage{comment}
\usepackage[color=yellow]{todonotes}

\title{Modular Design Patterns for Hybrid Actors}

\author{
    André Meyer-Vitali, Wico Mulder, Maaike H.T. de Boer\\
    TNO, Netherlands Organisation for Applied Scientific Research\\
    Anna van Buerenplein 1, 2595 DA, The Hague, The Netherlands\\
  
    \texttt{andre@meyer-vitali.com}, \texttt{wico.mulder@tno.nl}, \texttt{maaike.deboer@tno.nl}\\
}

\begin{document}

\maketitle

\begin{abstract}
Recently, a boxology (graphical language) with design patterns for hybrid AI was proposed, combining symbolic and sub-symbolic learning and reasoning. In this paper, we extend this boxology with actors and their interactions. The main contributions of this paper are: 1) an extension of the taxonomy to describe distributed hybrid AI systems with actors and interactions; and 2) showing examples using a few design patterns relevant in multi-agent systems and human-agent interaction.
\end{abstract}

\section{Introduction}
    \label{sec:introduction}

A boxology is a graphical language using boxes and (directed) lines, such as used in many types of diagrams. In earlier work, we presented a boxology \cite{vanbekkum2021} and stressed the use of design patterns to combine different paradigms in AI: knowledge-based reasoning and data-driven machine learning.
In this paper we extend on this and include the aspect of actors, along with the corresponding interaction with processes, instances and models. 

Design patterns can be visualised as a boxology, which is a representation of an organized structure in the form of a directed graph of labeled nodes, called boxes, and their relations and dependencies as arrows. Using a boxology improves communication and understanding of the structure of AI systems. This work aims to make complex distributed AI systems more transparent in terms of their design patterns. The modularity of the proposed design patterns also facilitates the design and engineering of complex AI systems. This allows for including descriptions of interactions among autonomous entities, such as software agents or robots. As a result, the boxology can be of relevance to a broader spectrum of distributed AI applications and topics such as multi-agent systems, federated or multi-agent learning, human-agent teaming and social intelligence.

An emerging field in AI is the combination of data-driven and knowledge-driven approaches \cite{huizing2020}. In addition, sharing of knowledge (collaborative learning and reasoning) and the collaboration between humans and actors remains a paramount challenge for truly hybrid AI. Extending the boxology with actors allows for specifying the aspects of human-machine interactions in such systems properly. We expect that the boxology approach may support discussions on autonomy, explainability, trustworthiness and sovereignty of AI systems. Furthermore, the inclusion of actors may help to address the responsibility and ownership of models and processes; topics that are being encountered on top of technical engineering questions - and of increasing importance in complex distributed AI systems. To the best of our knowledge no conceptual framework is available in which design elements can be discussed, compared, configured and combined with respect to those topics.

A framework describing a set of modular design patterns was initially developed by van Harmelen and ten Teije \cite{harmelen2019}, and extended by van Bekkum et al. \cite{vanbekkum2021}. In this paper, we use that conceptual framework as a basis (section 2) and extend it with the missing pieces, namely actors and interactions (section 3). In section 4, we show examples of using the design patterns with actors. We finish the paper with conclusions and directions for future work (section 5).

\subsection{Agency}
    \label{sec:agency}

Artificial Intelligence is based on the principles of autonomy and agency. Without them, intelligence cannot manifest itself. Agency is the capacity of individuals to act independently and to make their own free choices, according to Kant and others \cite{wikiAgency2021}. As mentioned above, (Artificial) Intelligence is unthinkable without agency. First of all, a minimal degree of autonomy is required to avoid purely predictable and reactive behaviour. Intelligence is, in fact, the result of a lot of interaction - be it in physiological (brains) or social systems (swarms) - and forms the basis and inspiration of most AI techniques (neural networks, knowledge graphs, communication networks). In the "Society of Mind" \cite{minsky1986} Minsky gave a number of examples for distributed AI and the importance of the interactive/social aspects of intelligence. Intelligence as a social and sociological phenomenon implies the formulation of intentional behaviour \cite{dennett1989}, where intents guide the planning of (inter-)actions. However, the control of social systems is not centralised, but shared among all participants, which leads to emergent behaviour of the system as a whole. As a result, individual autonomous actors need to consider and reflect upon the effect of their actions on other actors and their mental states, such as their knowledge, intentions, and beliefs (theory of mind \cite{premack1978} \cite{baron-cohen1991}), as well as their virtual or physical context (contextual embedding).

\section{Taxonomy \& Patterns}
    \label{sec:taxonomy}

\begin{figure}
\centering
    \includegraphics[width=\columnwidth]{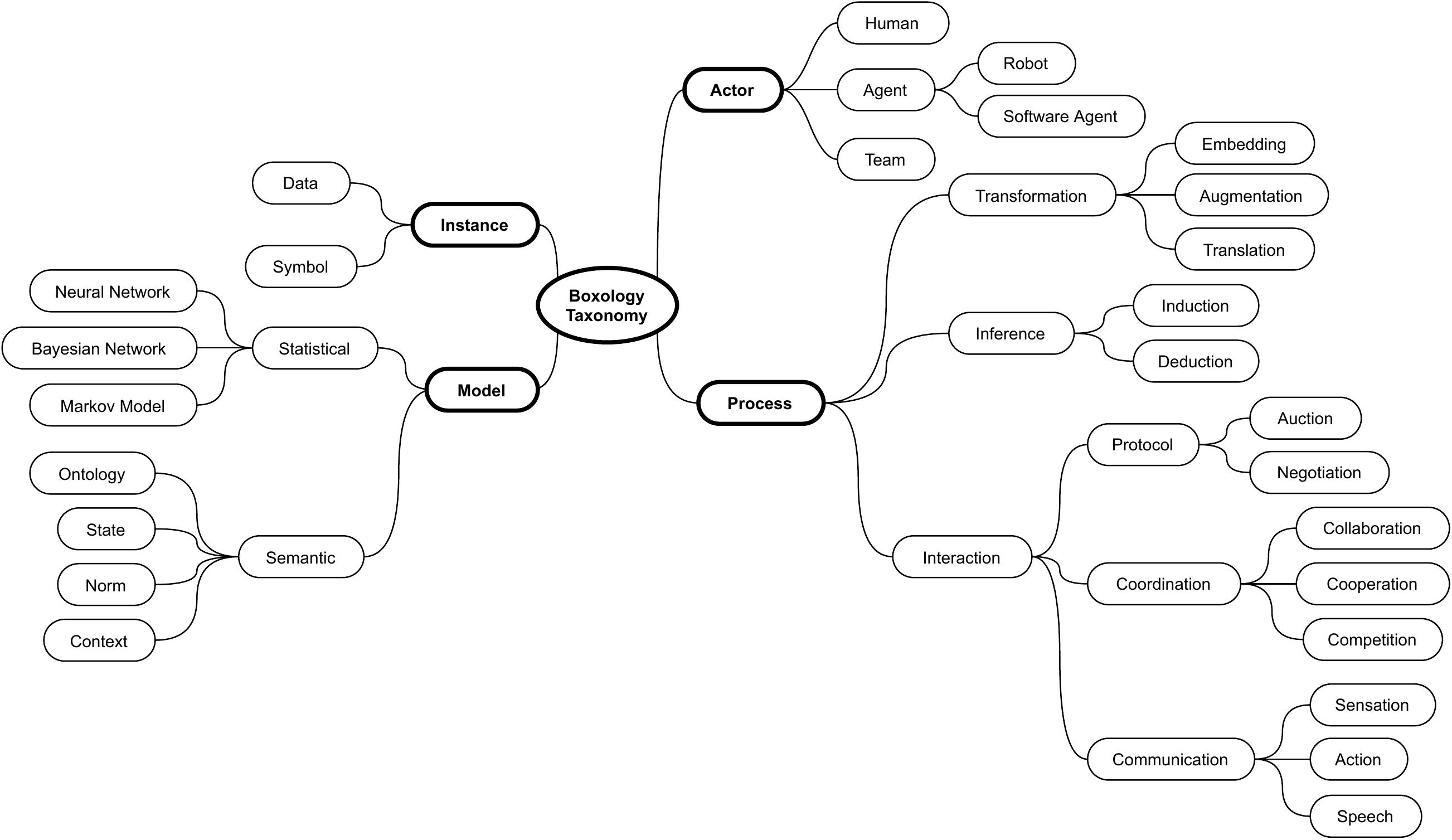}
    \caption{Taxonomy for Hybrid AI (simplified)}
    \label{fig:taxonomy}
\end{figure}

In order to describe design patterns of AI systems, a common terminology is required that defines a hierarchical taxonomy of processes and their inputs and outputs on various levels of abstraction. On the highest level of abstraction the concepts of \textit{Instance}, \textit{Model}, \textit{Process} and \textit{Actor} are defined (see figure \ref{fig:taxonomy}). The full taxonomy is larger and more detailed. For example, data and symbols, as well as induction and deduction, have a few subconcepts that are not shown here for the sake of clarity and space.

\textit{Instances} are the basic building blocks of “things”, examples or single occurrences of something. The two main classes of instances are data (numbers, texts, tensors, streams) and symbols (labels, relations, traces). In contrast to data, symbols must designate objects, classes or relations in the world. There must be compositional rules and a system of operations to generate new symbols \cite{Berkeley2008}.

\textit{Models} are descriptions of entities and their relationships. They are useful for inferring data and knowledge. Models in hybrid AI systems can be either (a) statistical models, such as (Deep) Neural Networks, Bayesian Networks and Markov Models, or (b) semantic models, such as Taxonomies, Ontologies and Knowledge Graphs.

In order to perform operations on instances and models, \textit{processes} define the steps that lead from inputs to results. The main types of processes are: (a) the transformation of instances and models, e.g. embedding in a vector space, and (b) inferencing, which includes induction or deduction. Induction is the generation of models either by means of generalising data using statistical machine learning algorithms or by means of knowledge engineering with domain experts. Deduction is the use or application of models and related instances for arriving at classifications, predictions and other kinds of conclusions, such as reasoning.

In section \ref{sec:extending} of this paper we work out the main concept \textit{Actor} and the related subconcepts added to processes (\textit{Interaction}), a few semantic models (\textit{State}, \textit{Norm} and \textit{Context}) and symbols (\textit{Request}, \textit{Reply}, etc.).

With the basic taxonomical terms in place, design patterns can be defined. The patterns can be used to describe different levels of abstraction: elementary patterns can be used in design descriptions, and more detailed patterns support architectural choices on types of models and instances used in an AI system.

\begin{figure}
\centering
    \includegraphics[width=0.7\columnwidth]{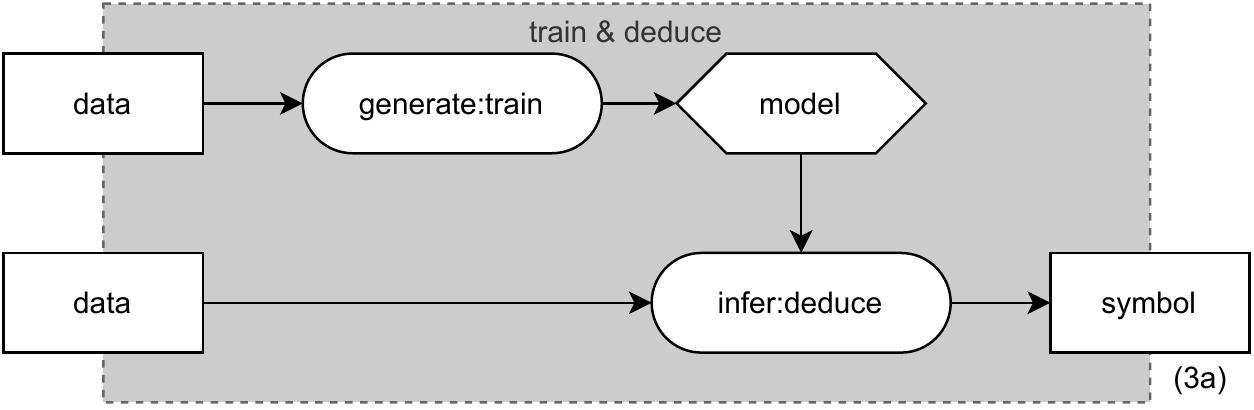}
    \caption{Machine Learning Pipeline}
    \label{fig:machinelearning}
\end{figure}

The main concepts and their subconcepts are visualised in diagrams using rectangles for instances, hexagons for models, ovals for processes and triangles for actors. The main category (as a verb) and the lowest subcategory in the hierarchy are used within the boxes separated with a colon, such as infer:deduce (where deduction is a subconcept of inference). Figure \ref{fig:machinelearning} describes the typical Machine Learning pipeline in terms of the taxonomy. This more complex pipeline (pattern 3a) consists of 2 elementary patterns (1a) and (2a) (not shown). Pattern (1a, top row in figure \ref{fig:machinelearning}) describes the standard training process, in which data is used to train a model (model generation). Pattern (2a) describes the application process, where (new) data is fed to the model and inferencing is performed to output a symbol, such as a classification. Hence, "Machine Learning" involves both an inductive step (generation by training) to produce a statistical model and a deductive step to use that model for classification or prediction.

\section{Extending the Boxology with Actors}
    \label{sec:extending}

The concept of actors was included in the previous version of the boxology. However, it was not yet worked out in detail, because the focus was on combining knowledge-driven reasoning with data-driven learning. 

Actors in a hybrid AI system are autonomous entities that can be either humans,  agents \cite{WoJe1995} \cite{wooldridge2009} \cite{weiss2013} or teams. 
Agents can take several forms, such as software agents (e.g., apps, services, mobile agents) or robots (physically embedded agents, e.g., drones or (parts of) autonomous vehicles). In the manufacturing domain actors can be cyber-physical systems, i.e., systems that consist of a physical part (e.g., a machine on the shopfloor) and a software part, often referred to as a \textit{digital twin} \cite{Steindl2020} \cite{Monostori2016}. In fact, there is renewed interest in distributed and multi-agent systems, be it in robotics or for the semantic web \cite{kirrane2021}, because they fit perfectly in current system architectures and infrastructures. Peer-to-peer networks (p2p), cloud services, edge computing, Internet of Things (IoT) and federated or multi-agent reasoning and learning (MARL) are perfect use cases for agent technology. They all make use of distributed systems with decentralised data and processing. In all these systems, coordination is necessary - either collaboratively or competitively - and agent technology has many proven solutions to these challenges. The actor concept, according to our taxonomy, includes robots (such as automated vehicles), software and human agents, as well as hybrid teams of all of them.

\subsection{Autonomous Actors}
    \label{sec:actors}

The relation between an actor and a process often comes with the notion of its role or responsibility. An actor can initiate a process or support an existing process.
Sometimes roles typically characterise an actor, e.g., when actors simply fulfill a particular designated task such as a robot, a data-provider in a data ecosystem, or a machine on a manufacturing shopfloor. Actors may also fulfill multiple roles to either the same or to different processes. Examples of such actors are digital twins or human machine operators.

Actors may interact with each other and form teams in order to coordinate tasks or learn to reach a common goal \cite{mulder2010}. Examples are swarms, multi-agent systems and human-agent teams. A team is a group of collaborating entities that can itself be regarded as a single actor \cite{talcott1951}. The aggregated team actors simplify the interaction models for many-to-many coordination by representing common goals and shared knowledge and resources (see below for examples).

Actors interact with each other using diverse (internal) behaviours and coordination protocols (subconcepts of \textit{Interaction}), such as collaboration, cooperation or competition \cite{Castelfranchi1998}. The interactions between actors define the so-called collective intelligence and may lead to emergent behaviour. As explained and demonstrated in the patterns below, team actors aggregate multiple concurrent lines of interaction concisely.

Autonomy is a gradual property \cite{vdvecht2007}, ranging from remotely controlled to selfish behaviour and all forms of pro-active behaviour in-between. It is subject to commitments made among agents: the more commitments an agent makes, the more it voluntarily limits its own autonomy temporarily. The rational and intentional scalability of agents' autonomous behaviour allows for various forms of coordination and team work. Intentionally and temporarily reducing its autonomy may well be in the best interest of the agent to achieve goals that cannot be achieved alone. Alternatively, a team can achieve more than each individual actor when actors consciously contribute to achieving the common team goal. This is when various forms of team coordination become essential.

\section{Applications}
    \label{sec:patterns}

In order to illustrate the above-mentioned concepts, three exemplary design patterns and a use case are explained below.

It is important to mention that the boxology allows to represent both the interactions among actors, as well as the internal structural patterns of AI systems. Consequently, it supports the design of hybrid AI systems using many different interaction, learning and reasoning patterns.

In order to visualise both aspects of interactive systems, a new design component was required for the diagrams. We chose for including a zooming function that is shown as a rectangular frame with a solid line and a light grey background (not to be confused with the patterns as shown with a dashed frame and dark grey background, see figure \ref{fig:machinelearning}). A badge at the top left of the frame indicates the type and instance of the component that is detailed inside the frame. Here, a triangular badge indicates that an actor is zoomed in. The same mechanism can also be used for zooming into processes, models or instances. In the case that an actor represents a team, it is possible to include individual actors within the frame as members of that team. This nesting is not suitable for individual actors, of course, but works for sub-processes, sub-models and sub-instances.

\subsection{Multi-Agent Learning and Reasoning}
    \label{sec:malr}

\begin{figure}
\centering
    \includegraphics[width=0.8\columnwidth]{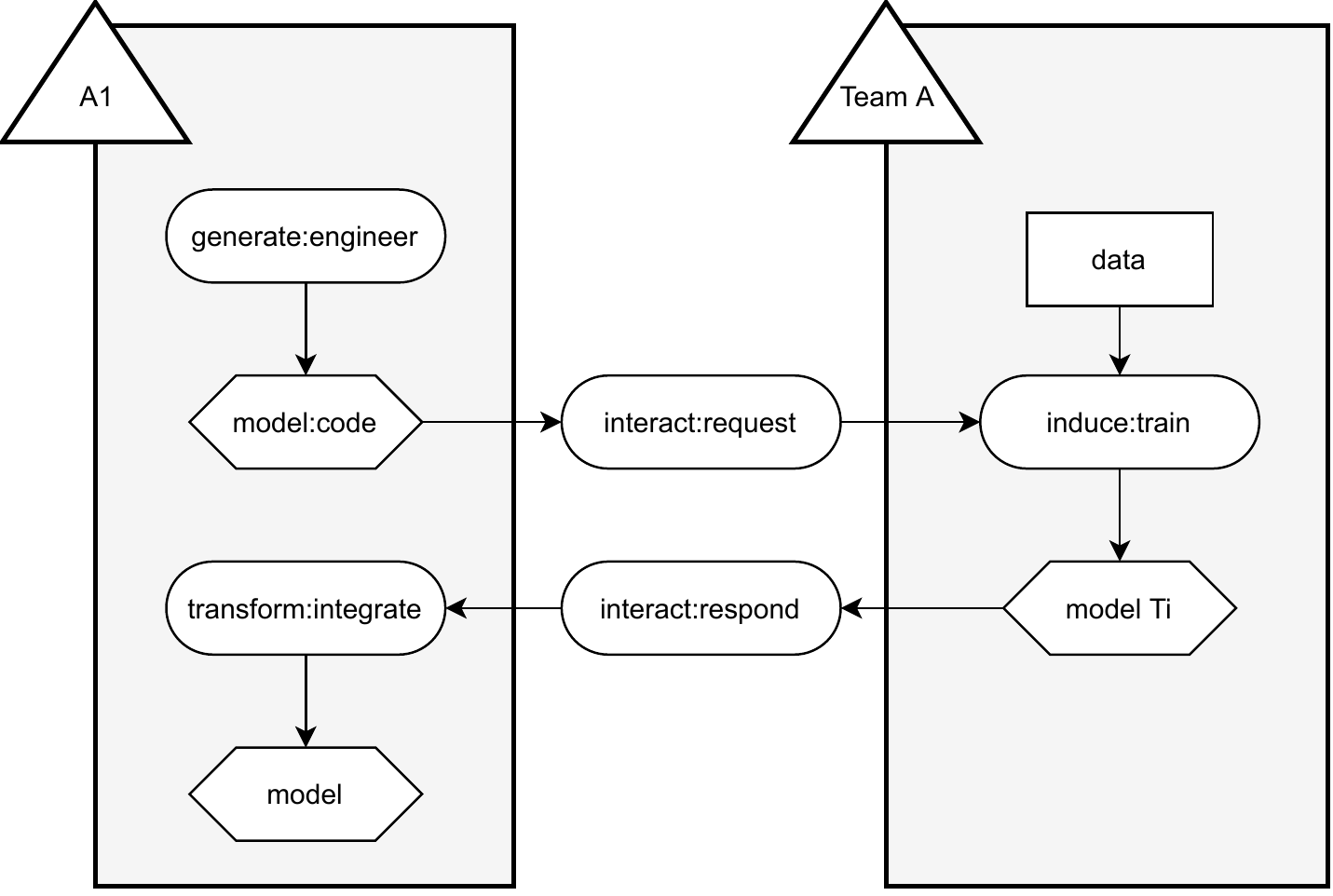}
    \caption{Multi-Agent Learning with Mobile Code}
    \label{fig:mobile}
\end{figure}

Interactive actors can collaborate for learning and reasoning and to share knowledge \cite{kairouz2019}. In figure \ref{fig:mobile} a multi-agent approach is shown where one actor requests a team to learn partial models using a provided learning algorithm, rather than collecting all data and processing it centrally. The learning algorithm is embedded in the requests (as a code model) and the partial model is returned in the responses. This approach is more efficient by allowing for parallel processing and it avoids privacy restrictions. In alternative approaches (not shown), an actor can simply send a request message and let the team members learn according to their own methods. Yet another approach is to use mobile agents that migrate to a target platform, process the data locally and move on with a trained model to other platforms. In all these patterns, partial models (models Ti) are integrated at some point by one designated party.

A related scenario is learning on demand, where an actor asks another actor for knowledge that the second one may already have learned and the requester needs at a certain point in time. For example, the second actor can teach the first one to perform a certain task.

Figure \ref{fig:mobile} shows two parties: Actor A1 and Team A. This is a very compact and concise representation of group behaviour. However, there can be more roles, e.g. a coordinator, that would be represented explicitly. Therefore, it is possible to represent more complex use cases where multiple roles can be used - eventually as subteams if necessary. In the case, as shown, each actor behaves according to the same interactive principles, thereby demonstrating a distributed interaction pattern without any central control (peer-to-peer).

\begin{figure}
\centering
    \includegraphics[width=1.0\columnwidth]{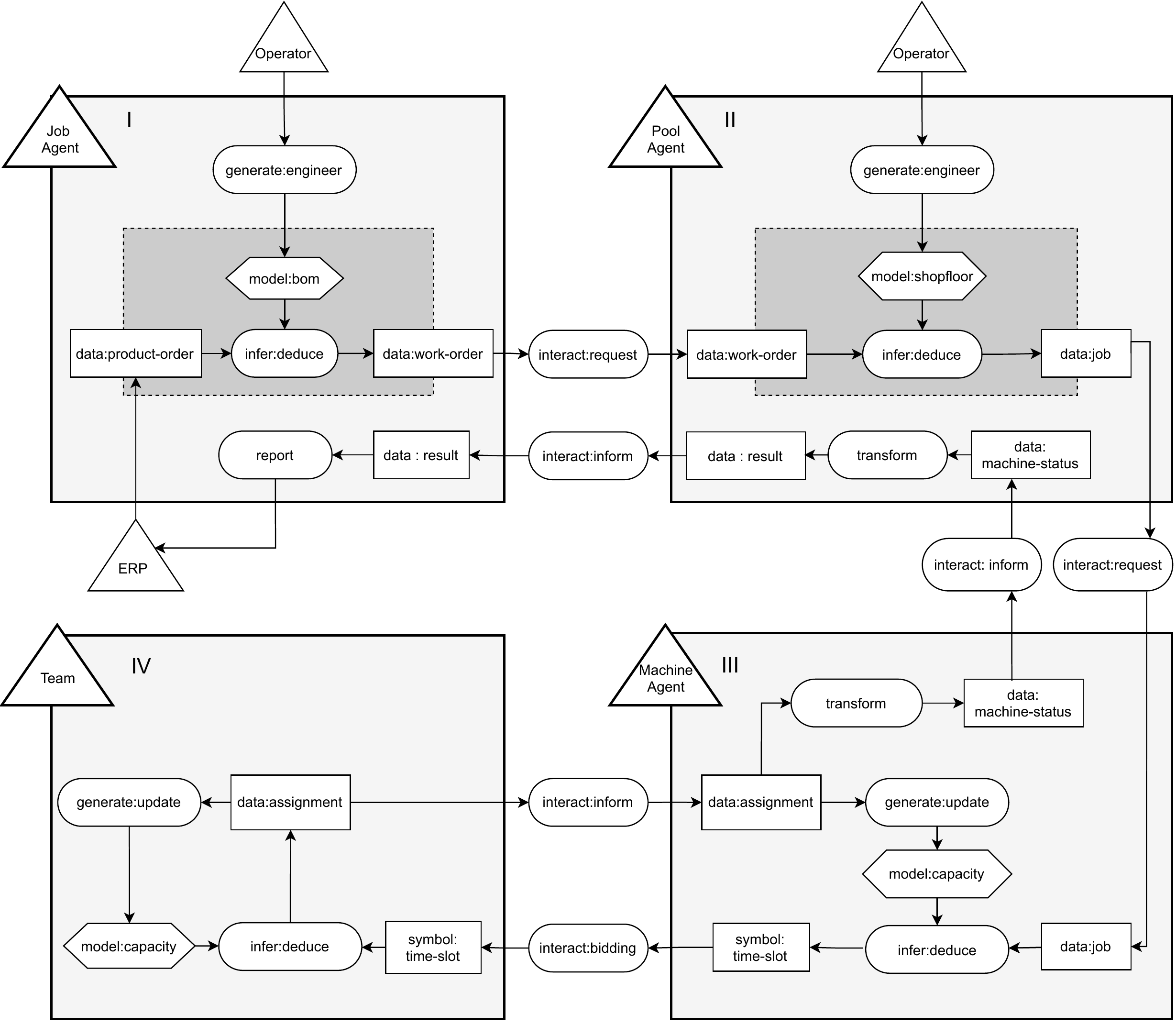}
    \caption{Distributed Planning using Job, Pool and Machine Agents}
    \label{fig:distributed_planning}
\end{figure}

\subsection{Use Case: Multi-Agent Planning}
    \label{sec:digiman}

In a manufacturing use case we use a similar approach for distributed planning. The boxology extends other engineering diagrams and focuses specifically on the underlying machine learning parts of the system.
This is shown in figure \ref{fig:distributed_planning}.

The distributed system consists of three types of software agents: job agents, pool agents and machine agents. The architecture consists of 4 parts. Part I shows that a Job Agent uses a bill of material model that was generated by a human operator. It takes product orders from an Enterprise Resource Planning (ERP) system and creates work orders. The work orders are sent to a pool agent, which controls a set of production machines. This is shown in part II. Note that the attention in the diagram is on the job agent, therefore the actor representing the job agent is placed in the corner of the grey box. The role of the operator is merely supportive here, as it generates a model of the factory shop floor. The pool agent receives requests for work orders. Using the model that was generated by a human operator, the job agent turns a work order into a specific job that can be carried out by one or more machines on the shop floor. Every eligible machine agent in the pool receives that job request and uses its own capacity model to judge whether it can perform the job in the requested time. If so, it informs the other machines in the pool about a possible time slot. This is shown in part III.  The interaction between the machines is a bidding process. The machine with the best option, i.e., that has offered highest bid, takes the job, updates its capacity model and informs the job agent. Part IV of the diagram shows the role of the team in that bidding process. The exact steps of this bidding process are not shown here. For the example, it suffices to show the main interaction between a machine agent and its team.

Eventually an incoming time slot leads via a bidding mechanism to the assignment of the job. The machine agent that will carry out the job will be informed, and the team (all the machine agents) updates the capacity model. Note, that part IV shows a composite actor (the team) in its interaction with a particular agent. The boxology is used to abstract from details such as the bidding. A process which often deserves its own details either in a boxology diagram (or a sub-diagram) or in a different type of interaction diagram as preferred by the system architect.

\subsection{Contextual Actors}
    \label{sec:context}

Situated or contextual actors are interacting with their physical or virtual environment. Various models exist for specifying the necessary steps and interactions for context-aware actors, such as BDI (beliefs, desires, intentions) \cite{rao1995} or OODA (observe, orient, decide, act) \cite{boyd2006}. For the interaction with their environment, actors perform communicative acts, namely sensing and acting. Actors can also perform communicative acts with other actors using language (speech acts, see section \ref{sec:negotiation}).

For contextual or situated actors, the concept of \textit{Deduction} is extended with \textit{Planning} and \textit{Interaction} with communicative acts  \cite{austin1962} \cite{searle1969}: \textit{Sensation}, \textit{Action} and \textit{Speech}. These subconcepts will be used in the patterns below. For certain interactions, additional symbols are required, such as \textit{Request}, \textit{Proposal}, \textit{Assignment}, or \textit{Result}. They play roles in interaction protocols, as defined by FIPA (see section \ref{sec:negotiation} below), and other communicative acts.

Relevant model-related concepts for actors are the semantic models of \textit{State}, \textit{Context} and \textit{Norm} \cite{dignum2004}. State includes reasoning components, such as beliefs, desires and intentions (to be used in BDI reasoning, see section \ref{sec:context} below). Context represents the external world (physical and virtual) that an actor can sense and act upon. An important aspect of the context (or environment) are resources. A \textit{Resource} model specifies the availability and capacity of objects and actors. For example, time limits or cognitive load could be relevant. \textit{Norms} define and regulate interaction among actors according to desirable and permitted behaviour (physical, legal, social or otherwise).

\begin{figure}
\centering
    \includegraphics[width=0.9\columnwidth]{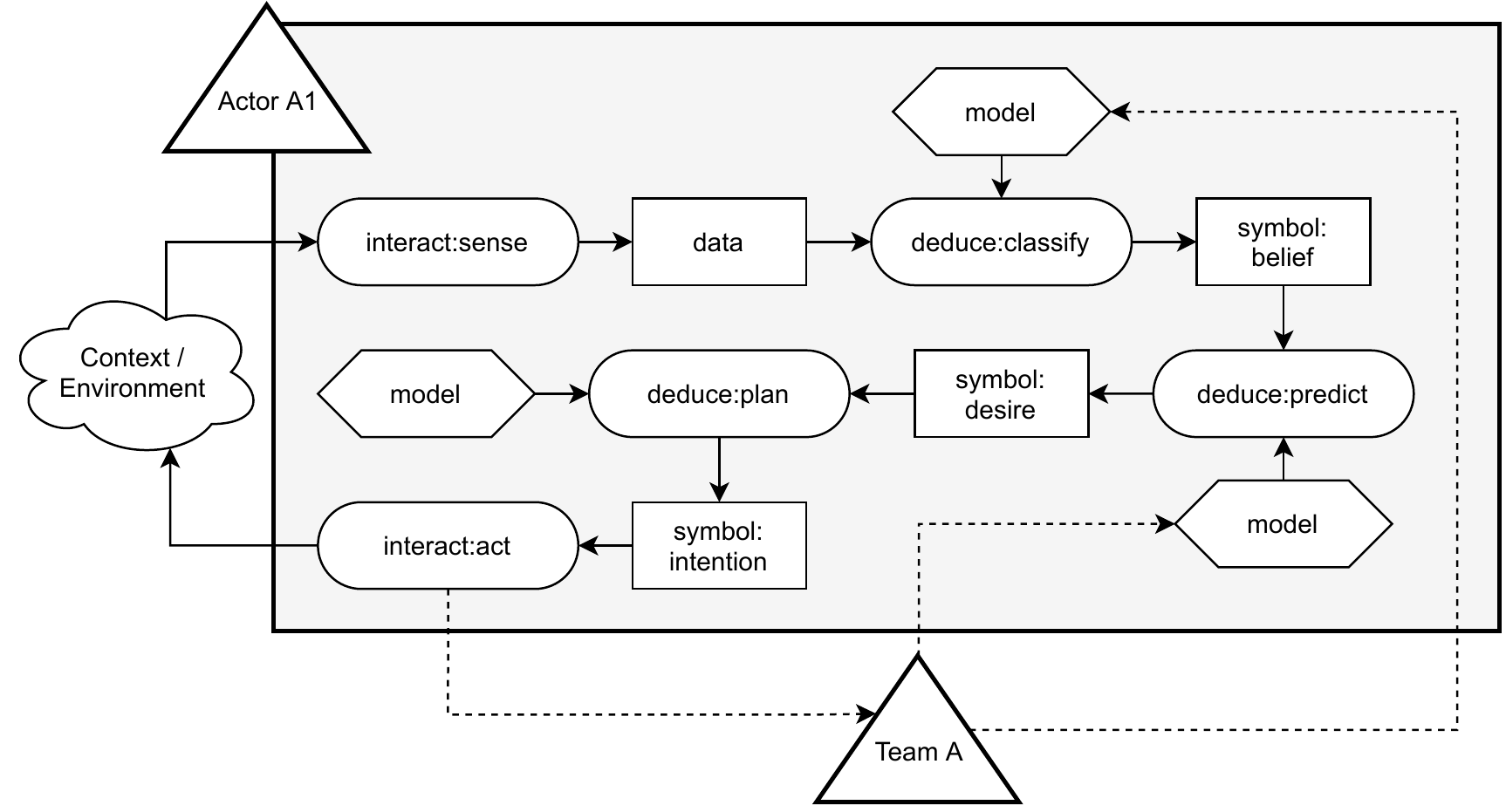}
    \caption{Contextual Agent using BDI}
    \label{fig:bdi}
\end{figure}

Figure \ref{fig:bdi} shows a contextually reasoning BDI actor. The actor senses data from the environment, classifies data to symbolic beliefs, predicts desires, plans intentions and acts on the environment. These processes make use of dedicated models, namely a world model (beliefs), a model of goals (desires) and a model of plans (intentions).

Besides acting on the environment, the actions can also take the form of speech acts to communicate with other actors of a team. This is indicated (without going into details) using the dotted arrows in the figure to show the incorporation of team knowledge that is shared via speech acts. Consequently, actors of a team can influence other actors' world and goal models (not without consent, of course).

\begin{figure}
\centering
    \includegraphics[width=1.0\columnwidth]{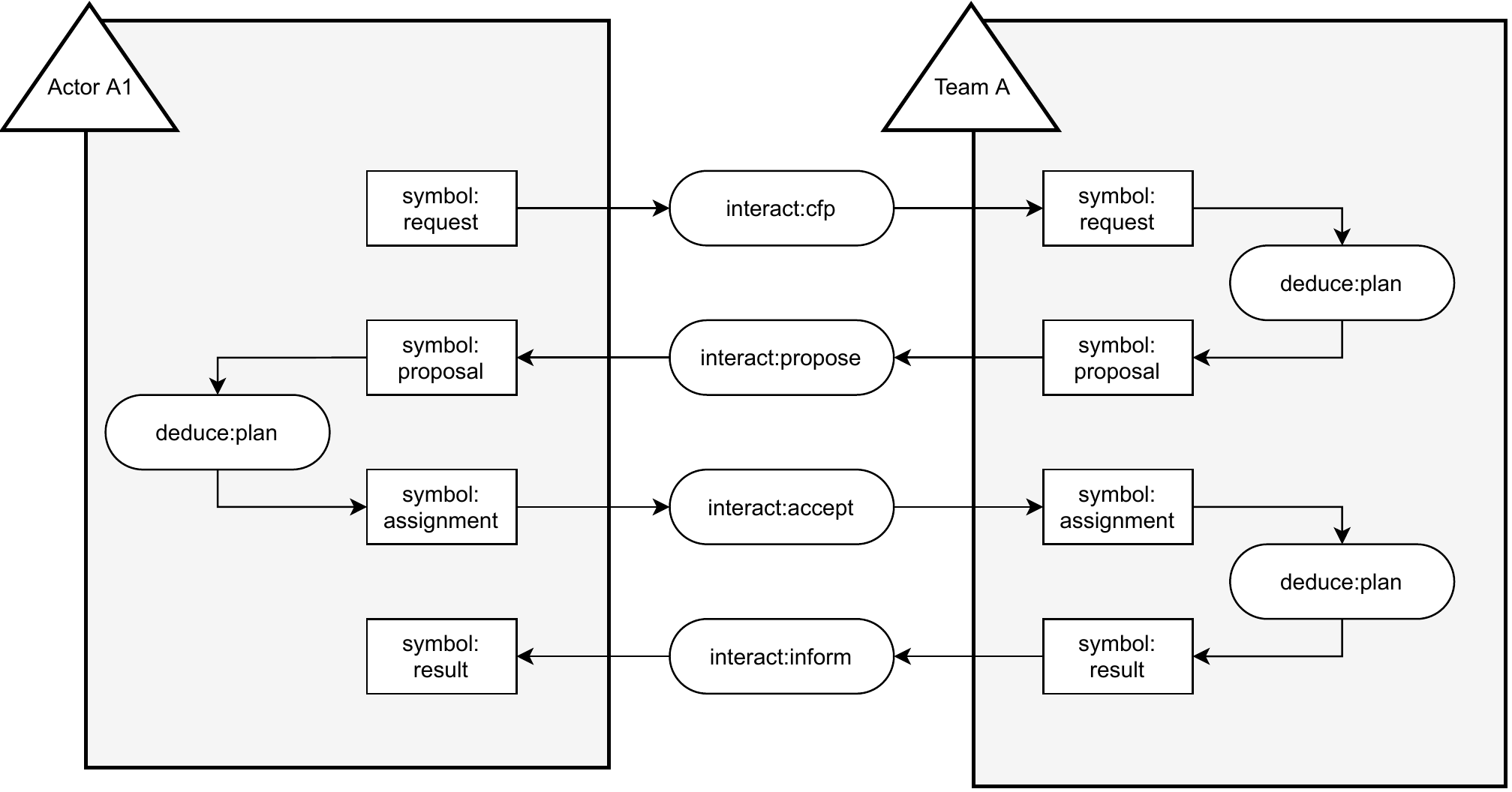}
    \caption{Team Negotiation with ContractNet}
    \label{fig:contractnet}
\end{figure}

\subsection{Negotiation}
    \label{sec:negotiation}

A well-known class of interactions are negotiations. As an example, figure \ref{fig:contractnet} shows a negotiation based on ContractNet \cite{poslad2000}, as specified by FIPA \cite{fipa-cn2002}. The planning steps of each agent (and team member) can be specified in more detail with more specific (induction and) deduction methods and corresponding models. Depending on the type of model learned in an inductive step, the deductive planning steps can use different strategies to come to concrete replies in a given situation.

The negotiation follows the above-mentioned interaction protocol and uses speech acts for the individual steps of the protocol. A call for proposals (CFP) is sent out by an actor to a team that it belongs to. Each team member may make a proposal (within a certain time limit) and the originator can accepts and reject them. When receiving an acceptance notification with an assignment, actors will produce a compliant result and return this to the requesting actor.

The pattern in figure \ref{fig:contractnet} does not specify the exact means of communication among the actors. Interaction, containing a CFP, for example, could occur via email, RMI, or any other message passing interface and infrastructure. However, the level of detail and abstraction can vary according to the maturity of the design. The goal is to define design patterns at a relatively high level of abstraction and to use a few of these for composing a hybrid AI system. Subsequently, the concepts can be made increasingly concrete. For example, choices will have to be made for the exact models and means of communication. From there, deployment of the system can be automated using off the shelf components.

\section{Conclusion and Future Work}
    \label{sec:conclusion}

In this paper, we extend the previous hybrid AI boxology with actors and interactions to make it usable in the field of distributed hybrid AI and human-agent interaction. We show three design patterns relevant in this field, where interaction among various actors is given shape and where actors learn and reason about their interactions. We also show a use case where agents interact in planning a production task. The visualisation of internal and external processes facilitates the identification and reuse of design patterns.

In future work, we plan to extend the taxonomy and the patterns even further, and to use the patterns for actually designing and efficiently implementing distributed hybrid AI systems. By avoiding a black box approach and explicitly describing the components and their interactions, such systems are expected to become more transparent, explainable and verifiable (predictable and reproducible). The extended boxology allows designers of distributed hybrid AI to produce more trustworthy systems.

\section*{Acknowledgement}
    \label{sec:acknowledgement}

This research was partially supported by TAILOR, a project funded by the EU Horizon 2020 research and innovation programme under GA No 952215.




\end{document}